\documentclass[11pt]{article}
\usepackage[]{EACL2023}
\usepackage{booktabs}
\usepackage{multirow}
\usepackage{times}
\usepackage{url}
\usepackage{latexsym}
\usepackage[utf8]{inputenc}
\usepackage[T5]{fontenc}
\usepackage{float}
\usepackage{adjustbox}
\usepackage{times}
\usepackage{latexsym}
\usepackage{microtype}
\usepackage{inconsolata}

\title{Revealing Weaknesses of Vietnamese Language Models Through Unanswerable Questions in Machine Reading Comprehension}

\author{Son Quoc Tran$^{1, 4}$, Phong Nguyen-Thuan Do$^{4}$, \\\textbf{Kiet Van Nguyen$^{2,3,4}$, Ngan Luu-Thuy Nguyen$^{2,3,4}$}\\
$^1$Denison University, Granville, OH, USA \\ 
$^2$University of Information Technology, Ho Chi Minh City, Vietnam\\
$^3$Vietnam National University, Ho Chi Minh City, Vietnam\\
$^4$The UIT NLP Group, Vietnam National University, Ho Chi Minh City\\
\texttt{tran\_s2@denison.edu, phongdntvn@gmail.com, \{kietnv, ngannlt\}@uit.edu.vn}}
\begin{document}
\maketitle
\begin{abstract}
Although the curse of multilinguality significantly restricts the language abilities of multilingual models in monolingual settings, researchers now still have to rely on multilingual models to develop state-of-the-art systems in Vietnamese Machine Reading Comprehension. This difficulty in researching is because of the limited number of high-quality works in developing Vietnamese language models. In order to encourage more work in this research field, we present a comprehensive analysis of language weaknesses and strengths of current Vietnamese monolingual models using the downstream task of Machine Reading Comprehension. From the analysis results, we suggest new directions for developing Vietnamese language models. Besides this main contribution, we also successfully reveal the existence of artifacts in Vietnamese Machine Reading Comprehension benchmarks and suggest an urgent need for new high-quality benchmarks to track the progress of Vietnamese Machine Reading Comprehension. Moreover, we also introduced a minor but valuable modification to the process of annotating unanswerable questions for Machine Reading Comprehension from previous work. Our proposed modification helps improve the quality of unanswerable questions to a higher level of difficulty for Machine Reading Comprehension systems to solve.
\end{abstract}
\section{Introduction}
Machine Reading Comprehension (MRC) is a challenging research field in Natural Language Processing, in which systems learn to predict answers for the questions inputted by users given a relevant context. MRC has many real-world applications such as Open Domain Question Answering \cite{chen-etal-2017-reading} and conversational Question Answering \cite{reddy-etal-2019-coqa}. Thanks to the rapid development of pre-trained large language models, performances of MRC systems show substantial progress. Pre-trained large language models are typically deep learning models designed based on the architecture of the Transformers model \cite{NIPS2017_3f5ee243}. These models are pre-trained on very large text corpora using unsupervised tasks such as Masked Language Model and Next Sentence Prediction \cite{devlin-etal-2019-bert}. After the pre-training phase, researchers can leverage the language understanding of these models by fine-tuning them on downstream tasks such as MRC. After being fine-tuned, these language models can achieve state-of-the-art performances on many benchmarks.

Researchers also pre-train multilingual models which are transformers-based models pre-trained with text corpora in over 100 languages \citet{conneau-etal-2020-unsupervised,devlin-etal-2019-bert}. Although multilingual models do not rely on direct cross-lingual supervision while being pre-trained, they can achieve surprisingly high performances on different tasks in multilingual settings. Besides, these multilingual models also excel in monolingual settings, especially in low-resource languages, where the number of high-quality works in developing monolingual language models is still limited. However, the abilities of multilingual language models are restricted by the curse of multilinguality \cite{conneau-etal-2020-unsupervised}: pre-training a multilingual model with a fixed capacity on an increasing number of languages only improves its performances up to a certain point. Therefore, pre-trained multilingual models often show many language weaknesses compared to monolingual counterparts in monolingual settings.

Following the success of pre-trained models in English \cite{devlin-etal-2019-bert,zhuang-etal-2021-robustly}, researchers all over the world carry out many high-quality works in pre-training monolingual language models such as CamemBERT \cite{chan-etal-2020-germans} in French, GELECTRA \cite{martin-etal-2020-camembert} in German, and PhoBERT \cite{nguyen-tuan-nguyen-2020-phobert} in Vietnamese. These monolingual models also achieve state-of-the-art performance on numerous benchmarks, directly empowering the field of Natural Language Processing to develop in their respective languages.

Facilitated by the development of pre-trained language models, MRC has recently also shown great progress in many languages. For example, RoBERTa \cite{roberta}, CamemBERT \cite{martin-etal-2020-camembert} and GELECTRA \cite{chan-etal-2020-germans} achieve near human performances on SQuAD \cite{rajpurkar-etal-2018-know}, FQuAD \cite{dhoffschmidt-etal-2020-fquad,fquad20} and GermanQuAD \cite{moller-etal-2021-germanquad}, respectively. However, for other low-resource languages, such as Vietnamese, the performances of pre-trained language models are significant far lower than that of humans \cite{viquad20}. We can explain these difficulties in research by the underdevelopment of Vietnamese monolingual language models. As a result, most researchers \cite{vireader,vlspmrc1,vlspmrc2,Nguyen_2020} in Vietnamese MRC have to use multilingual models, which have many limitations in monolingual settings, as the cores of their MRC systems. 

The difficulties that Vietnamese MRC researchers encounter, together with the limited number of works on Vietnamese monolingual models, suggest that more high-quality research into Vietnamese monolingual models is urgently needed. Therefore, in order to suggest new directions for these future works, we attempt to reveal the language weaknesses of monolingual models by analyzing the performances of monolingual models in comparison with those of multilingual ones.

In this work, we choose to investigate the performances of models on MRC because it is a suitable task for exploring the weaknesses of language models from multiple linguistic aspects. MRC allows us to examine the performance of models on lexical aspects, single-sentence level aspects, and multi-sentence level aspects of natural language. For instance, in order to answer "Who" questions, MRC models must be competent in recognizing the person's name in a sentence, demonstrating their proficiency in Named Entity Recognition. Besides, to fully understand the given context, MRC models are expected to acquire extraordinary Reading Comprehension skills such as coreference resolution and bridging, which are part of the multi-sentence level aspects of language understanding.

We focus our analysis on unanswerable questions because unanswerable questions proposed by \citet{viquad20} are much more challenging than answerable questions in the same dataset, which directly creates more materials for us to reveal the language weaknesses of models. Additionally, since \citet{viquad20} proposed a novel method for annotating unanswerable questions, which involves instructing annotators to use various techniques to transform answerable questions into unanswerable ones instead of generating unanswerable questions from scratch, UIT-ViQuAD 2.0 has successfully introduced many new types of unanswerable questions. Therefore, we have a more diverse range of language aspects to analyze the performances of models on.

we initially examine the performance of monolingual and multilingual models on the UIT-ViQuAD 2.0 development set. However, we concern that the development set of UIT-ViQuAD 2.0 may not be sufficiently challenging to expose the language weaknesses of models on specific language aspects. Hence, we annotate a new set of high-quality unanswerable questions on an out-of-domain corpus to further analyze the language proficiency of both monolingual and multilingual models.

Our contributions are summed as follows:
\begin{enumerate}
    \item Our work successfully discovers different language weaknesses and strengths of Vietnamese monolingual models. Results from our work provide good directions for future works on more robust Vietnamese monolingual models.
    \item To more accurately assess the language abilities of models, we propose a new method for annotating high-quality unanswerable questions that successfully further challenge current systems in MRC. 
    \item Results from our analysis reveal that new high-quality Vietnamese Machine Reading Comprehension benchmarks are urgently needed.
\end{enumerate}

\section{Related Work}
\textbf{Unanswerable Questions. } Unanswerable questions in MRC draw much attention from the research community after the publication of SQuAD 2.0 \cite{rajpurkar-etal-2018-know}. Following the guidelines proposed by \citet{rajpurkar-etal-2018-know}, unanswerable questions in MRC are introduced in MRC of other languages such as French in FQuAD 2.0 \cite{fquad20} and Vietnamese in UIT-ViQuAD 2.0 \cite{viquad20}. The research community commonly refers to unanswerable questions in SQuAD, FQuAD, and UIT-ViQuAD as "artificial unanswerable questions" because annotators are instructed to intentionally create questions that cannot be answered using the information provided in the given context. On the other hand, unanswerable questions that naturally arise are also introduced recently in Natural Questions \cite{kwiatkowski-etal-2019-natural} and TyDi QA \cite{clark-etal-2020-tydi}, in which the evidence documents are provided after the questions are written by annotators.\\
\textbf{Multilingual versus Monolingual Models. } \citet{vulic-etal-2020-probing} probe an empirical analysis on monolingual BERTs and mBERT across six languages and five different lexical tasks. They show that Monolingual BERT encodes significantly more lexical information than mBERT.

Besides, \citet{rust-etal-2021-good} compare pre-trained multilingual language models with monolingual counterparts regarding their monolingual task performances in nine languages and five tasks to reveal the reason for the gap between the performances of monolingual models and multilingual models. This comprehensive analysis later reveals that while pre-training data size played a vital role in the performances of language models on downstream tasks, the monolingual tokenizers designed by native speakers are also an important reason for the high performances of monolingual models in single-language settings. Results from this analysis show that \citet{nguyen-tuan-nguyen-2020-phobert} significantly contributed to the development of Vietnamese language models with a high-quality tokenizer that is suitable for the unique linguistic features of Vietnamese.
\begin{table*}[h]
\centering
\begin{tabular}{@{}llllcc@{}}
\toprule
 &          & EM(\%)    & F1(\%)   & Recall\textsubscript{unanswerable}(\%)  & Recall\textsubscript{answerable}(\%)  \\ \cmidrule(l){2-6} 
\multirow{2}{*}{monolingual}  & WikiBERT & 46.51 & 55.84 & 50.68 & 74.37 \\
 & PhoBERT     & 63.52          & 75.87          & 73.37          & \textbf{89.21} \\ \cmidrule(l){2-6} 
\multirow{3}{*}{multilingual} & mBERT\textsubscript{our}    & 57.66 & 66.84 & 65.84 & 80.47        \\
& mBERT\textsubscript{VLSP}    & 53.55 & 63.03 & - & -        \\
 & XLM-RoBERTa & \textbf{67.84} & \textbf{78.15} & \textbf{75.86} & 88.81 \\ \bottomrule
\end{tabular}
\caption{Performance of models on the UIT-ViQuAD 2.0 Development set}
\label{overall-performane}
\end{table*}
\section{Models and Analysis Method}
\subsection{Models}
In this work, to highlight the weaknesses of Vietnamese language models, we compare the performances of two Vietnamese monolingual language models with those of two multilingual language models.\\
\textbf{Multilingual Language Models.} We choose mBERT \cite{devlin-etal-2019-bert} and XLM-RoBERTa \cite{conneau-etal-2020-unsupervised} as two multilingual models. Because we are investigating the weaknesses of existing models of each language model type, we decide to use XLM-RoBERTa\textsubscript{LARGE}, which outperforms XLM-RoBERTa\textsubscript{BASE} in almost all tasks of natural language processing. XLM-RoBERTa\textsubscript{LARGE} has 24 transformer-based layers with 560M parameters and was trained on 2394.3 GiB of text in 100 languages, in which 137.3 GiB of 24.7 billion word tokens is Vietnamese text. On the other hand, mBERT has 12 transformer-based layers with 178M parameters and was trained in 104 languages, including Vietnamese.\\
\textbf{Monolingual Language Models. }We choose the large version of PhoBERT \cite{nguyen-tuan-nguyen-2020-phobert}, and Vietnamese WikiBERT \cite{pyysalo-etal-2021-wikibert} as two competitive monolingual models against multilingual counterparts. PhoBERT\textsubscript{LARGE} is a transformer-based model with 370M parameters and is trained with 20GiB of 3 billion Vietnamese word tokens. The critical difference of PhoBERT from multilingual models is that PhoBERT segments Vietnamese words before applying the Byte-Pair encoding methods \cite{sennrich-etal-2016-neural} to the pre-training data. For example, while multilingual models tokenize the word ``học sinh''(\textit{student}) as two tokens, ``học'' and ``sinh'', PhoBERT treats this whole word as a single token ``học\_sinh'' This is because white space in Vietnamese is used to separate the syllables instead of words.

On the other hand, Vietnamese WikiBERT has 101M parameters and is trained with 172M Vietnamese tokens. Because researchers developing Vietnamese WikiBERT are not Vietnamese native speakers, they do not acknowledge the unique linguistic features of the Vietnamese language as \citet{nguyen-tuan-nguyen-2020-phobert} do.

In this paper, for simplicity, we will refer to PhoBERT\textsubscript{LARGE}, XLM-RoBERTa\textsubscript{LARGE} and Vietnamese WikiBERT as PhoBERT, XLM-RoBERTa, and WikiBERT, respectively.
\subsection{Analysis Method}
Following previous works \cite{rajpurkar-etal-2016-squad,rajpurkar-etal-2018-know,nguyen-etal-2020-vietnamese}, we use two metrics, Exact Match (EM) and F1-score, to evaluate the overall performances of different models on Reading Comprehension task.
\begin{itemize}
    \item \textbf{EM}: (Exact Match) The percentage of answers predicted by the MRC system match exactly any one of the gold answer(s) annotated by the human reader.
    \item \textbf{F1}: F1-score measured the average overlap between predicted answers with those in the gold answers. For each question, we calculate the F1 score of predicted answer with each gold answer, and take the maximum F1 as the F1 of the corresponding question. 

\end{itemize}
Because we carry out our analysis on the test set that requires models having abilities to recognize unanswerable questions, we also take into consideration the performances of models in classifying answerable and unanswerable questions. Performances on classification tasks are reported in our analysis as Recall on answerable questions and unanswerable questions.
\begin{itemize}
    \item \textbf{Recall\textsubscript{unanswerable}}: The percentage of unanswerable questions that the model correctly predicts as not having the answer in the given context.
    \item \textbf{Recall\textsubscript{answerable}}: The percentage of answerable questions that model attempt to answer. In order to focus on the classification task, this metric does not consider whether the model predicts the correct answer.
\end{itemize}
Then, in order to analyze the performances of models on different language aspects, we annotate each unanswerable question into one of 7 unanswerable types, most of which are inspired by \cite{viquad20}.

Besides, as we focus on suggesting new directions for works in developing Vietnamese monolingual models, instead of pointing out the weaknesses of any single model, we focus on determining different hard language aspects that are challenging for all investigated Vietnamese language models. Thus, we define two new concepts for this purpose:
\begin{itemize}
    \item \textbf{Monolingual hard unanswerable questions}: Unanswerable questions that both WikiBERT and PhoBERT attempt to answer.
    \item \textbf{Multilingual hard unanswerable questions}: Unanswerable questions that both mBERT and XLM-RoBERTa attempt to answer.
\end{itemize}
These concepts of monolingual and multilingual hard unanswerable questions empower us to focus on the language weaknesses that both monolingual models have compared to the weaknesses of both mBERT and XLM-RoBERTa. Thus, we can encourage future research to follow effective methods from previous works and develop new methods to deal with the existing weaknesses. To compare the results between different experiments, we calculate the percentage of monolingual and multilingual hard unanswerable questions over the total number of unanswerable questions in each unanswerable type.
\subsection{Experimental Settings}
All models are trained with 28,457 questions in training set of UIT-ViQuAD 2.0 \cite{viquad20} in 2 epochs. We use Adam optimizer \cite{DBLP:journals/corr/KingmaB14} with learning rate of $2\cdot10^{-5}$, $\beta_1 = 0.9$, $\beta_2 = 0.999$, and 100 warm-up steps for all 4 models. We fine-tuned all four models on a single NVIDIA Tesla K80 provided by Google Colaboratory. Due to these limited resources in computation, we have to fine-tune our models with a small number of samples per batch. The fine-tuning batch size we use for XLM-RoBERTa, mBERT, WikiBERT is 4, while 8 is the batch size in fine-tuning PhoBERT. We then evaluate models on the development set of UIT-ViQuAD 2.0 in Section 4 and Parallel UIT-VinewsQA in Section 5. 
\section{Analysis on UIT-ViQuAD 2.0}
\begin{table*}[h]
\resizebox{\textwidth}{!}{%
\centering
\begin{tabular}{cccc}
\hline
\multicolumn{1}{l}{}            & \begin{tabular}[c]{@{}c@{}} \# Unanswerable questions\\in developement set\end{tabular}          & \begin{tabular}[c]{@{}c@{}}Monolingual \\ hard unanswerable questions (\%)\end{tabular} & \begin{tabular}[c]{@{}c@{}}Multilingual\\ hard unanswerable questions(\%)\end{tabular} \\ \hline
Antonym                         & 80                        & \textbf{15.00}                                                                                 & 16.25                                                                                 \\
Overstatement \& Understatement & 68                        & \textbf{8.82}                                                                                  & 14.71                                                                                 \\
Entity Swap                     & 360                       & 14.17                                                                                 & \textbf{6.39}                                                                                 \\
Normal Word Swap                & 383                       & \textbf{15.67}                                                                                 & 16.97                                                                                 \\
Relation Reverse                & 138                       & 28.99                                                                                 & \textbf{13.04}                                                                                 \\
Adverbial Clause Swap           & 21                        & 38.10                                                                                  & \textbf{33.33}                                                                                  \\
Modifiers Swap                 & 91                        & \textbf{13.19}                                                                                 & 19.78                                                                                 \\
Dataset Noise                   & 27                        & 40.74                                                                                 & 33.33                                                                                  \\ \hline
Total                           & 1,168 & 17.20                                                            & \textbf{14.00}                                                            \\ \hline
\end{tabular}%
}
\caption{Number of monolingual and multilingual hard unanswerable questions alongside with the number of unanswerable questions in the the full development set by types}
\label{tab:hard-unanswerable-result}
\end{table*}
\subsection{Overall Performance}
Table \ref{overall-performane} shows the performance of models on the development set of UIT-ViQuAD 2.0 \cite{viquad20}. XLM-RoBERTa outperforms other three models on EM, F1 and Recall\textsubscript{unanswerable} while slightly underperforms PhoBERT on Recall\textsubscript{answerable}.

The development set of UIT-ViQuAD 2.0 was used as the public test for VLSP2021: Machine Reading Comprehension \cite{viquad20}. Based on the results published after the shared task, our fine-tuned mBERT substantially outperforms the mBERT baseline of the organizers. 
\subsection{Performance on Unanswerable Questions}
We then analyze the performances of models on different unanswerable types of unanswerable questions (see Table \ref{tab:categories} in \ref{sec:appendix1} for examples). We closely follow unanswerable question types defined by \citet{viquad20}. However, based on our observation, when using \textit{Entity Swap} for creating unanswerable questions, annotators might unintentionally reverse the relation of entities in the original questions. Therefore, in order to exploit these important questions for revealing language weaknesses of monolingual models, we define \textit{Relation Reverse} as a new unanswerable type for our analysis and analyze it separately from \textit{Entity Swap} type. Results from our analysis (Table \ref{tab:hard-unanswerable-result}) show that questions of \textit{Relation Reverse} type are much more challenging for models than those of \textit{Entity Swap} type.

Results from our analysis successfully reveal some language weaknesses of monolingual models. As reported in Table \ref{tab:hard-unanswerable-result}, the performances of Vietnamese monolingual models on \textit{Entity Swap} and \textit{Relation Reverse} types are significantly lower than those of multilingual models. This result shows us that the ability to represent the relationships between different entities in the context of Vietnamese monolingual models are significantly inferior than multilingual models.

However, monolingual models show strong performances on \textit{Modifiers Swap type} which requires language models to have a good ability in understanding the modified relationships between different words in the sentence. In other words, Vietnamese monolingual models acquire a better ability in low-level lexical and grammatical features of Vietnamese than multilingual counterparts do. We hypothesize that the unusual characteristics of the Vietnamese language  pose significant challenges for multilingual models. If an adjective is used as a noun modifier in Vietnamese, the adjective must go after the main noun instead of before, as in English and many other resource-rich languages.

On the other hand, monolingual and multilingual models show little difference in their performances on unanswerable question types of \textit{Antonym}, \textit{Overstatement \& Understatement}, and \textit{Adverbial Clause Swap}. However, we are concerned that the number of high-quality unanswerable questions of those types in the development set of UIT-ViQuAD 2.0 is not enough to reveal weaknesses of language models in these aspects of language. Therefore, we annotate a new small high-quality benchmark on the corpus of UIT-VinewsQA \citep{van2020new}, which is another high-quality Vietnamese MRC dataset. 

\section{Analysis on Parallel UIT-VinewsQA}
\begin{table}[!h]
\resizebox{\linewidth}{!}{
\centering
\begin{tabular}{@{}lllll@{}}
\toprule
             & \# entities & \# paragraphs & \# sentences & \# tokens \\ \midrule
UIT-VinewsQA & 4,465       & 500           & 8,131        & 159,857   \\
UIT-ViQuAD   & 6,476       & 557           & 3,208        & 78,628    \\ \bottomrule
\end{tabular}
}
\caption{Number of entities, paragraphs, sentences and tokens of UIT-VinewsQA and UIT-ViQuAD development sets predicted by Trankit, a light-weight Transformer-based toolkit for multilingual natural language processing \cite{nguyen-etal-2021-trankit}}
\label{tab: language-vinews-viquad}
\end{table}
\begin{table*}[h]
\resizebox{\textwidth}{!}{
\centering
\begin{tabular}{@{}llllccc@{}}
\toprule
 &          & EM(\%)    & F1(\%)   & F1\textsubscript{answerable}(\%) & Recall\textsubscript{unanswerable}(\%)  & Recall\textsubscript{answerable}(\%)   \\ \cmidrule(l){2-7} 
\multirow{2}{*}{monolingual}  &WikiBERT & 36.61 & 48.12 &61.61 & 34.64 & 81.79 \\
 & PhoBERT     & 40.54         & 56.86     &80.16    & 33.57          & \textbf{95.71}\\ \cmidrule(l){2-7} 
\multirow{2}{*}{multilingual} &    mBERT    & 41.61 & 52.35 & 65.06& 38.64 & 83.21       \\
 &  XLM-RoBERTa & \textbf{49.64} & \textbf{62.50} & \textbf{81.80} & \textbf{43.21} & 95.00\\ \bottomrule
\end{tabular}
}
\caption{Performances of models on Parallel UIT-VinewsQA}
\label{vinewsqa-performane}
\end{table*}
UIT-VinewsQA is an extractive question answering dataset on Vietnamese healthcare news articles, most of which are narrative articles instead of informative like articles on the Wikipedia platform. Moreover, healthcare articles in UIT-VinewsQA are written for people with different education levels, so the sentence structure used in these articles must be simpler than that of Wikipedia articles. Therefore, as presented in Table \ref{tab: language-vinews-viquad}, UIT-VinewsQA has some linguistic differences from UIT-ViQuAD, such as
\begin{itemize}
    \item UIT-VinewsQA has fewer entities per sentence than UIT-ViQuAD. This significantly reduces the challenging level of recognizing relations between entities in the given context of extractive question answering task. Therefore, unanswerable questions of types such as \textit{Entity Swap} and \textit{Relation Reverse} are not as challenging for language models in UIT-VinewsQA compared to UIT-ViQuAD 2.0.
    \item UIT-VinewsQA has fewer tokens per sentence than UIT-ViQuAD, which leads to simpler sentence structures across the corpus. 
\end{itemize}

\subsection{Benchmark Annotations}
When annotating new unanswerable questions on UIT-VinewsQA, we strictly follow the procedure proposed by \citet{viquad20}: we transform answerable questions extracted from the development set of UIT-VinewsQA into unanswerable questions. However, to promote the diversity of unanswerable questions, we intentionally sample our answerable questions based on their reasoning skills inspired by \citet{van2020new} (word matching, paraphrasing, single-sentence reasoning, multiple-sentence reasoning). For each answerable reasoning skill - unanswerable question type pair, we annotated ten unanswerable questions. Therefore, we have a benchmark of 280 unanswerable questions of four answerable reasoning skills and seven unanswerable question types in addition to 280 answerable questions extracted from the UIT-VinewsQA development set. We name this benchmark Parallel UIT-VinewsQA because each answerable question in the benchmark is accompanied by a corresponding unanswerable question.

Besides, during the annotating process, we do not show our annotators the answers to the original (answerable) questions and ask them to annotate answers for these questions before transforming original questions into unanswerable ones. We only include an unanswerable question into our benchmark if the annotator correctly answers the corresponding answerable question. This helps us strictly require our annotators to grasp a ``big picture'' of the given context instead of merely focusing on the sentences containing answers to the original questions. In later analysis, we find out that this process significantly improves the quality of questions of all unanswerable types.
\subsection{Performance on Parallel UIT-VinewsQA}
\begin{table*}[h]
\resizebox{\textwidth}{!}{
\centering
\begin{tabular}{cccc}
\hline
\multicolumn{1}{l}{}            & \begin{tabular}[c]{@{}c@{}}\# Unanswerable questions\\in Parallel UIT-VinewsQA  \end{tabular}        & \begin{tabular}[c]{@{}c@{}}Monolingual \\ hard unanswerable questions (\%)\end{tabular} & \begin{tabular}[c]{@{}c@{}}Multilingual\\ hard unanswerable questions (\%)\end{tabular} \\ \hline
Antonym                         & 40                        & 55.00                                                                                 & \textbf{37.50}                                                                                 \\
Overstatement \& Understatement & 40                        & 50.00                                                                                  & \textbf{45.00}                                                                                 \\
Entity Swap                     & 40                       & 25.00                                                                                 & \textbf{22.50}                                                                                 \\
Normal Word Swap                & 40                       & 45.00                                                                                 & \textbf{35.00}                                                                                 \\
Relation Reverse                & 40                       & \textbf{35.00}                                                                                 & 37.50                                                                                 \\
Adverbial Clause Swap           & 40                        & 72.50                                                                                  & \textbf{62.50}                                                                                  \\
Modifiers Swap                 & 40                        & \textbf{55.00}                                                                                 & 57.50                                                                                 \\
Total                           & 280 & 48.21                                                            & \textbf{42.50}                                                            \\ \hline
\end{tabular}

}
\caption{Number of monolingual and multilingual hard unanswerable questions alongside with the number of unanswerable questions in the Parallel UIT-VinewsQA by types}
\label{tab:vinewsqa-hard-unanswerable-result}
\end{table*}

Table \ref{vinewsqa-performane} show that all considered models  achieve only from $33.57\%$ to $43.21\%$ on Recall\textsubscript{answerable} when evaluated on the 280 newly annotated unanswerable questions. his cannot be attributed solely to the out-of-domain context, as the models performed well on the 280 answerable questions extracted from UIT-ViNewsQA, achieving the highest F1 score of 81.80\% among the four models. This result indicates that while UIT-VinewsQA is considered one of the high-quality Vietnamese MRC datasets, it does not fully reveal the existing weaknesses of MRC systems.

Our newly generated unanswerable questions thus give us much more materials to analyze the weaknesses and strengths of monolingual models in MRC. We then analyze the weaknesses of language models by examining the percentage of monolingual and multilingual hard unanswerable questions out of the total number of unanswerable questions. 

Due to the linguistic features of the UIT-VinewsQA corpus shown in Table \ref{tab: language-vinews-viquad}, \textit{Entity Swap} and \textit{Relation Reverse} types of unanswerable questions are no longer challenging as they are in UIT-ViQuAD. On the other hand, the most notable result from our analysis is that \textit{Antonym} type is significantly more challenging for monolingual models than for multilingual models. As the unanswerable questions of \textit{Antonym} type in SQuAD 2.0 \cite{nguyen-etal-2020-vietnamese} often require language models good lexical knowledge to correctly recognize, monolingual models are believed to have advantages over multilingual counterparts. This is because \cite{vulic-etal-2020-probing} show that monolingual models often encode significantly more lexical information than monolingual models. However, because we are following the process of annotating unanswerable questions proposed by \citet{viquad20} on a different corpus, we hypothesize that there may be some significant changes in unanswerable questions of \textit{Antonym} type in our benchmark.
\subsection{Analysis on Antonym Type}
Closely examining the performances of models on each unanswerable question of \textit{Antonym} type, we see that monolingual models often fail to recognize an unanswerable question when the antonym used to create that question does not explicitly contradict the context. Based on this observation, we believe that these questions should be analyzed separately from other questions of \textit{Antonym} type to understand the language weaknesses of monolingual models fully. We then divide \textit{Antonym} type into two new types of \textit{Implicit Antonym} and \textit{Explicit Antonym} to further explore the effects each type have on two types of language models (see Figure \ref{implcit-explicit-example} in \ref{sec:appendix2} for examples). In short, language models can correctly predict unanswerable questions of \textit{Explicit Antonym} using only lexical knowledge. However, to recognize an unanswerable question of \textit{Implicit Antonym}, models must acquire an adequate amount of high-level semantic knowledge.

\begin{table*}[h]
\centering
\begin{tabular}{@{}cccc@{}}
\toprule
 &
  Full Benchmark &
  \begin{tabular}[c]{@{}c@{}}Hard Monolingual \\ Unanswerable questions (\%)\end{tabular} &
  \begin{tabular}[c]{@{}c@{}}Hard Multilingual\\ Unanswerable questions (\%)\end{tabular} \\ \midrule
Explicit Antonym &
  25 &
  40.00 &
  \textbf{32.00} \\
Implicit Antonym &
  15 &
  80.00 &
  \textbf{46.67} \\ \bottomrule
\end{tabular}
\caption{Number of monolingual and multilingual hard unanswerable questions alongside with the number of unanswerable questions in the Parallel UIT-VinewsQA in Implicit and Explicit Antonym types}
\label{tab:explicit-implicit}
\end{table*}
Our analysis (Table \ref{tab:explicit-implicit}) reveals that while monolingual models show comparable performance on \textit{Explicit Antonym} type to multilingual models, \textit{Implicit Antonym} type is significantly more challenging for monolingual models than for multilingual models. This result proves that monolingual models lack skills in representing the relations between context and the adjective describing the context, which is part of high-level semantic knowledge.
\section{Conclusion}
In this paper, we present the first comprehensive analysis to reveal the weaknesses of state-of-the-art Vietnamese language models. Our experiments show that while Vietnamese language models demonstrate good lexical and grammatical abilities in Vietnamese, they show inferior performances when questions require high-level semantic knowledge to successfully identify the unanswerability. This general result from our analysis shows that the inferior performances of Vietnamese language models on Machine Reading Comprehension task are mainly due to its inferior ability in grasping the big ``picture'' of the given context.

Besides, our analysis also show that Vietnamese MRC benchmarks overestimate the comprehension skills of models in some language aspects, so  state-of-the-art performances on MRC benchmarks does not accurately reflect the progress of Vietnamese Machine Reading Comprehension.

\section{Future Directions}
Based on the results from our analysis, we suggest several future directions for both Vietnamese monolingual language models and Vietnamese MRC benchmarks.
\subsection{Language Models}
Our analysis shows that monolingual models, especially PhoBERT, acquire comparable abilities in recognizing the differences in lexical information between unanswerable questions and the given context. However, monolingual models show poor performances when encountering unanswerable questions that require the ability to comprehend a bigger ``picture''. For example, while monolingual models perform very well on unanswerable questions that use explicit antonyms, they often have difficulties in recognizing unanswerable questions when these questions are created using implicit antonyms. We explain this phenomenon by the findings of \citet{zhang-etal-2021-need} as pre-training language models on larger text copora results in significant improvement on downstream tasks that require high-level semantic and factual knowledge such as Machine Reading Comprehension. Therefore, when encountering unanswerable questions that require ability to grasp big ``picture,'' models pre-trained with smaller text corpora will show lower performances. Hence, the small size of pre-training corpora of PhoBERT and WikiBERT may be the main reason for their poor performances in MRC.

Scaling the pre-training data size of PhoBERT will further develop this model and empower it to achieve state-of-the-art performances on different benchmarks of Machine Reading Comprehension. Besides, we believe that introducing a new unsupervised task for the pre-training phase that focuses on improving the high-level semantic and factual knowledge of pre-trained models also plays an integral role in developing language models in the future.
\subsection{Benchmarks}
\textbf{Unanswerable Questions. } Although UIT-ViQuAD 2.0 successfully further introduced new types of artificially unanswerable questions, our work in Section 5 shows that current unanswerable questions in the development test of UIT-ViQuAD 2.0 are still not challenging enough. In order to increase the challenging levels of unanswerable questions, we believe that more high-quality works on adversarial human annotation for unanswerable questions are needed. These works can follow the guidelines of adversarial human annotation for answerable questions \cite{bartolo-etal-2020-beat}. Results of these works can reveal different techniques to annotate hard unanswerable questions and therefore be valuable for improving the guidelines for unanswerable questions annotation for Machine Reading Comprehension.\\
\textbf{Quality of Benchmark. } On the other hand, as we have shown in section 5, although PhoBERT and XLM-RoBERTa achieve high performances on the UIT-VinewsQA development set, our unanswerable questions reveal that these two models do not truly understand the context to give the correct answer for questions in the original development set. We hypothesize that questions in UIT-VinewsQA enable machine reading comprehension systems with shortcut learning knowledge \cite{lai-etal-2021-machine} to achieve high performance due to biases in annotating process. Therefore, we believe that studies on how Vietnamese machine reading comprehension systems are currently evaluated are important for tracking the progress of Vietnamese language systems.

\bibliography{anthology,custom}
\bibliographystyle{acl_natbib}
\newpage
\appendix

\section{Supplementary material}
\subsection{Unanswerable Types Examples}
\label{sec:appendix1}
Table \ref{tab:categories} shows examples of the unanswerable types that we focus our analysis on. Most unanswerable types in our work are inspired by the original work of \citet{viquad20}. 
\begin{table*}[]
\resizebox{\textwidth}{!}{
\begin{tabular}{p{3cm}ll}

\hline
\textbf{Reasoning}    & \textbf{Description}                                                                                                   & \multicolumn{1}{c}{\textbf{Example}}                                                                                      \\ \hline
Antonym               & Antonym used                                                                                                           & \begin{tabular}{p{14cm}}\textbf{Sentence}: Vào năm 1171, Richard khởi hành đến Aquitaine với mẹ mình và Henry phong ông là Công tước xứ Aquitaine theo yêu cầu của Eleanor.  (\textit{In 1171, Richard departed to Aquitaine with his mother and Henry, who had appointed him as the Duke of Aquitaine at the request of Eleanor.})\\ 
\textbf{Original question}: Richard khởi hành đến Aquitaine với mẹ vào năm nào?  (\textit{In what year did Richard depart to Aquitaine with his mother?}) \\ textbf{Unanswerable question}: Richard khởi hành từ Aquitaine với mẹ vào năm nào?  (\textit{In what year did Richard depart from Aquitaine with his mother?}) \end{tabular}                                                                                                                                                                                                                                                                                                                                                                                                                                                                                                                                    
\\ \hline
Overstatement         & \begin{tabular}{p{5cm}} Word that has similar meaning but with a higher shades of meaning is used\end{tabular}   & \begin{tabular}{p{14cm}}\textbf{Sentence}: Ngày 9 tháng 11 năm 1989, vài đoạn của Bức tường Berlin bị phá vỡ, lần đầu tiên hàng ngàn người Đông Đức vượt qua chạy vào Tây Berlin và Tây Đức. (\textit{On November 9, 1989, several parts of the Berlin Wall were collapsed, and for the first time thousands of East Germans crossed into West Berlin and West Germany.)}\\  \textbf{Original question}: Bức tường Berlin đã bị sụp đổ một vài đoạn vào ngày nào? (\textit{On which date were some parts of Berlin Wall collapsed?})\\ \textbf{Unanswerable question}: Bức tường Berlin đã bị sụp đổ hoàn toàn vào ngày nào? (\textit{On which date was Berlin Wall completely collapsed?})\end{tabular}                                                                                                                                                                                                                                                                                                                                                                                                                                                                                   \\ \hline
Understatement        & \begin{tabular}{p{5cm}}Word that has similar meaning but with a lower shades of meaning is used\end{tabular}     & \begin{tabular}{p{14cm}}\textbf{Sentence}: Quân đội Nhật Bản chiếm đóng Quảng Châu từ năm 1938 đến 1945 trong chiến tranh thế giới thứ hai. (\textit{The Japanese army occupied Guangzhou from 1938 to 1945 during the second world war.})\\  \textbf{Original question}: Khi Chiến tranh Thế giới thứ hai xảy ra thì Quảng Châu bị nước nào chiếm đóng? (\textit{During the World War II, Guanzong was occupied by which country?})\\ \textbf{Unanswerable question}: Khi Chiến tranh Thế giới thứ hai xảy ra thì Quảng Châu bị nước nào đe dọa? (\textit{During the World War II, Guanzong was attacked by which country?})\end{tabular}                                                                                                                                                                                                                                                                                                                                                                                                                                                                                                                                                \\ \hline
Entity Swap           & Entity replaced by other entity                                                                                        & \begin{tabular}{p{14cm}}\textbf{Sentence}: Là cảng Trung Quốc duy nhất có thể tiếp cận được với hầu hết các thương nhân nước ngoài, thành phố này đã rơi vào tay người Anh trong chiến tranh nha phiến lần thứ nhất. (\textit{As the only Chinese port accessible to most foreign merchants, the city fell to the British during the First Opium War.})\\  \textbf{Original question}: Trong cuộc chiến nào thì Anh Quốc đã chiếm được Quảng Châu? (\textit{In which war did Britain occupy Guangzhou?})\\ \textbf{Unanswerable quetion}: Trong cuộc chiến nào thì Nhật đã chiếm được Quảng Châu? (\textit{In which war did Japan occupy Guangzhou?})\end{tabular}                                                                                                                                                                                                                                                                                                                                                                                                                                                                                                                     \\ \hline
Relation Reverse      & \begin{tabular}{p{5cm}} Reverse the relation between two entities\end{tabular}        & \begin{tabular}{p{14cm}}\textbf{Sentence}: Một lần nữa, Gandhi bị bắt giam, và chính quyền tìm cách đập tan ảnh hưởng của ông bằng cách cách li hoàn toàn ông và các người đi theo ủng hộ. (\textit{Once again, Gandhi was imprisoned, and the government sought to crush his influence by completely isolating him from his followers.})\\\textbf{Original question}: Chính quyền làm cách nào để đập tan ảnh hưởng của Gandhi?(\textit{How does the government crush Gandhi's influence?})\\ \textbf{Unanswerable quetion}: Gandhi làm cách nào để đập tan ảnh hưởng của Chính quyền?(\textit{How does Gandhi crush the influence of the government?})\end{tabular} \\ \hline
Normal Word Swap      & \begin{tabular}{p{5cm}}A normal word replaced by another normal word\end{tabular}                                & \begin{tabular}{p{14cm}}\textbf{Sentence}: Sự phát hiện của Hofmeister năm 1851 về các thay đổi xảy ra trong túi phôi của thực vật có hoa {[}...{]} (\textit{Hofmeister's discovery in 1851 of changes occurring in the embryo sac of flowering plants {[}...{]}})\\ \textbf{Original question}: Năm 1851 nhà sinh học Hofmeister đã tìm ra điều gì ở thực vật có hoa? (\textit{What did the biologist Hofmeister discover in flowering plants in 1851?})\\ \textbf{Unanswerable question}: Năm 1851 nhà sinh học Hofmeister đã công nhận điều gì ở thực vật có hoa? (\textit{What did the biologist Hofmeister accept in flowering plants in 1851?})\end{tabular}                                                                                                                                                                                                                                                                                                                                                                                                                                                                                                                                    \\ \hline
Adverbial Clause Swap & \begin{tabular}{p{5cm}}Adverbial clause replaced by another adverbial clause related to the context\end{tabular} & \begin{tabular}{p{14cm}}\textbf{Sentence}: Trước đó Phạm Văn Đồng từng giữ chức vụ Thủ tướng Chính phủ Việt Nam Dân chủ Cộng hòa từ năm 1955 đến năm 1976. Ông là vị Thủ tướng Việt Nam tại vị lâu nhất (1955–1987). Ông là học trò, cộng sự của Chủ tịch Hồ Chí Minh. (\textit{Pham Van Dong previously held the position of Prime Minister of the Democratic Republic of Vietnam from 1955 to 1976. He was the longest-serving Prime Minister of Vietnam (1955-1987). He was a student and collaborator of President Ho Chi Minh.})\\ \textbf{Original question}: Giai đoạn năm 1955-1976, Phạm Văn Đồng nắm giữ chức vụ gì? (\textit{What position did Pham Van Dong hold during the period from 1955 to 1976?}) \\ \textbf{Unanswerable question}: Khi là cộng sự của chủ tịch Hồ Chí Minh, Phạm Văn Đồng nắm giữ chức vụ gì? (\textit{As a collaborator of President Ho Chi Minh, what position did Pham Van Dong hold?})\end{tabular}                                                                                                                                                                                                                                                     \\ \hline
Modifiers Swap      & \begin{tabular}{p{5cm}} Modifier of one word in the given context is used for another word\end{tabular}        & \begin{tabular}{p{14cm}}\textbf{Sentence}: Các phần mềm giáo dục đầu tiên trong lĩnh vực giáo dục đại học (cao đẳng) và tập trung được thiết kế chạy trên  máy tính đơn (hoặc các thiết bị cầm tay). Lịch sử của các phần mềm này được tóm tắt trong SCORM 2004 2nd edition Overview (phần 1.3) (\textit{The first educational software in the field of higher education (college) and concentration was designed to run on a single computer (or portable devices). The history of these software is summarized in SCORM 2004 2nd edition Overview (section 1.3).}) \\  \textbf{Original question}: Lịch sử của các phần mềm giáo dục đầu tiên trong lĩnh vực giáo dục đại học (cao đẳng) được tóm tắt, ghi nhận ở đâu? (\textit{Where did the history of the first educational software in the field of higher education} (\textit{college}) \textit{was summarized and recorded?}) \\ \textbf{Unanswerable quetion}: Lịch sử của các phần mềm giáo dục trong lĩnh vực giáo dục đại học (cao đẳng) được tóm tắt, ghi nhận đầu tiên ở đâu? (\textit{Where did the history of the educational software in the field of higher education (college) was first summarized and recorded?})\end{tabular} \\ \hline

\end{tabular}
}
\caption{Categories of unanswerable questions in UIT-ViQuAD 2.0. Most of categories are inspired by and adopted from \citet{viquad20}}
\label{tab:categories}
\end{table*}

\subsection{Implicit and Explicit Antonym}
\label{sec:appendix2}
\begin{figure*}[ht]
\centering
\includegraphics[width=\textwidth]{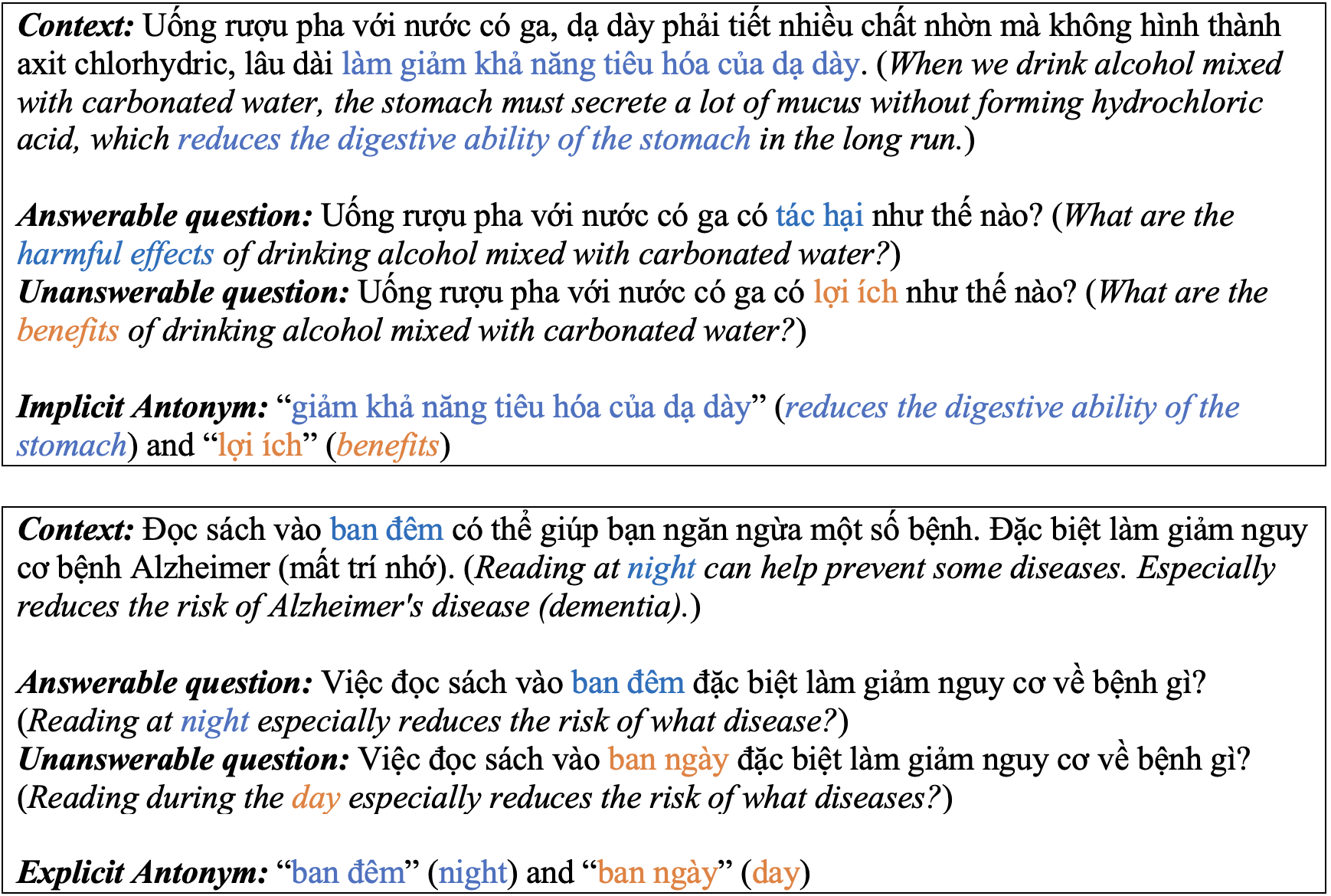}
\caption{Example of Implicit and Explicit Antonym}
\label{implcit-explicit-example}
\end{figure*}

Figure \ref{implcit-explicit-example} shows examples for Implicit Antonym and Explicit Antonym, which are defined in Section 5 of our analysis.

\end{document}